# Adaptive Class Emergence Training: Enhancing Neural Network Stability and Generalization through Progressive Target Evolution

Jaouad DABOUNOU[1*]

[1*] Professor, Hassan First University of Settat, Faculté Sciences et Techniques, Department of Mathematics Informatics and Engineering Science, Settat, Morocco

*Corresponding author E-mail: jaouad.dabounou@uhp.ac.ma;

**Abstract**

Recent advancements in artificial intelligence, particularly deep neural networks, have pushed the boundaries of what is achievable in complex tasks. Traditional methods for training neural networks in classification problems often rely on static target outputs, such as one-hot encoded vectors, which can lead to unstable optimization and difficulties in handling non-linearities within data. In this paper, we propose a novel training methodology that progressively evolves the target outputs from a null vector to one-hot encoded vectors throughout the training process. This gradual transition allows the network to adapt more smoothly to the increasing complexity of the classification task, maintaining an equilibrium state that reduces the risk of overfitting and enhances generalization. Our approach, inspired by concepts from structural equilibrium in finite element analysis, has been validated through extensive experiments on both synthetic and real-world datasets. The results demonstrate that our method achieves faster convergence, improved accuracy, and better generalization, especially in scenarios with high data complexity and noise. This progressive training framework offers a robust alternative to classical methods, opening new perspectives for more efficient and stable neural network training.

## 1. Introduction

Neural networks, especially deep neural networks, have achieved remarkable success in the field of artificial intelligence in recent years, enabling significant advancements in numerous complex tasks [1-4]. However, despite this progress, optimizing the training processes of these networks remains a major challenge [5-8]. Conventional training approaches, particularly for classification tasks, often rely on methods that can be suboptimal in terms of efficiency and stability.

In this paper, we propose an innovative approach to neural network training, inspired by the principles of dynamic equilibrium observed in physical systems [9,10]. Our method, which we call "Adaptive Class Emergence Training" (ACET), introduces two key concepts: progressive evolution of target outputs and equilibrium-based optimization.

ACET fundamentally alters the way neural networks learn to classify. Instead of imposing one-hot encoding target vectors from the outset, our approach progressively evolves these vectors, starting from a uniform distribution and gradually reaching the final one-hot encodings. This progression allows the network to adapt more smoothly and naturally to the increasing complexity of the classification task.

Concurrently, we introduce a dynamic equilibrium mechanism that regulates the optimization process. Network weight updates are performed only when the system significantly deviates from the defined equilibrium state, thereby reducing superfluous computations and stabilizing the learning process.

Our experiments, conducted on various synthetic and real-world datasets, including the MNIST [11], and Melanoma Skin Cancer Dataset demonstrate the effectiveness of ACET. The results reveal notable improvements in terms of convergence speed, final accuracy, and training stability compared to classical methods. On the MNIST dataset, for instance, our approach achieved an accuracy of 99.51%, surpassing the conventional method which typically doesn't attain 99.30%, while significantly reducing training time. Same results hold for Melanoma Skin Cancer Dataset.

This study opens new perspectives for neural network optimization, proposing a learning paradigm that aligns more closely with the principles of progressive adaptation observed in natural systems [12]. The implications of this approach extend beyond mere performance improvement, offering a new conceptual framework for understanding neural network learning.

Our findings suggest that ACET could have broad applications in improving neural network training across various domains, particularly where precise classification and generalization are crucial.

## 2. Theoretical Foundations

In this section, we present the theoretical foundations of the proposed approach for training neural networks with progressive target output updates and equilibrium-based weight adjustments. The approach is grounded in two key concepts: equilibrium-based training and progressive output updates.

### 1.1. Equilibrium-Based Training

We formalize the concept of equilibrium in neural networks as a state where the network's output closely aligns with the target outputs, such that the difference between them is minimal. Let $f(x; \theta)$ denote a neural network with parameters $\theta$ and input data $x$, and $y^*$ be the target output associated with $x$. We define the network to be in equilibrium if the loss function satisfies:

$$L(f(x; \theta), y^*) \leq \varepsilon \tag{1}$$

where $\varepsilon$ is a small positive threshold representing the acceptable level of error or disequilibrium. This differs from traditional approaches where weight updates occur after every forward pass, regardless of the network's proximity to the target output.

### 1.2. Regularity of the Progressive Target Output Update Function

We introduce a framework for progressively updating target outputs during training. We define the intermediate target output at time $t \in [0, 1]$ as:

$$y_c(t) = t \cdot y_c^* + (1 - t)\frac{1}{n_{classes}} \mathbb{1} \tag{2}$$

where $y_c^*$ is the final one-hot encoded target vector, $n_{classes}$ is the number of classes, and $\mathbb{1}$ is a vector of ones of size $n_{classes}$. The function $y_c(t)$ as a linear combination of $t$ and constant terms, is continuous and differentiable on [0,1]. The derivative of $y_c(t)$ with respect to $t$ is:

$$\frac{dy_c(t)}{dt} = y_c^* - \frac{1}{n_{classes}} \mathbb{1} \tag{3}$$

Obviously $n_{classes} \geq 2$, the norm of the derivative verifies:

$$\left\| \frac{dy_c(t)}{dt} \right\| \leq K, \text{ where } K = 1 - \frac{1}{n_{classes}} < 1 \tag{4}$$

Hence, the function $y_c(t)$ is continuously differentiable on [0,1] with a bounded derivative. This demonstrates regularity of the progressive target output update function.

Futhermore, the gradient of the cross-entropy loss with respect to the evolving target $\frac{\partial L}{\partial y_c}$ is bounded. Indeed, we have

$$L(f(x;\theta), y_c(t)) = \sum_{n=1}^{n_{classes}} y_c(t)_n \log(f(x;\theta)_n)$$

And

$$\left( \frac{\partial L}{\partial y_c(t)} \right)_n = -\log(f(x;\theta)_n)$$

$$f(x;\theta)_n = \frac{e^{y(t)_n}}{\sum_{i=1}^{n_{classes}} e^{y(t)_i}}$$

Given that $f(x;\theta)_n$ is bounded away from 0 (due to the softmax function), the logarithm of the probabilities is bounded. Therefore, there exists a constant $C$ such that:

$$\left\| \frac{\partial L}{\partial y_c(t)} \right\| \leq C \tag{5}$$

Relations (4) and (5) give:

$$\left\| \frac{\partial L}{\partial y_c} \right\| = \left\| \left( \frac{\partial L}{\partial y_c(t)} \right)^T \frac{dy_c(t)}{dt} \right\| \leq C \tag{6}$$

### 2.1. Dynamic Equilibrium

**Theorem 1**: Network Equilibrium with Progressive Target Updates

Let $f(x;\theta)$ be a neural network with parameters $\theta$ and input $x$, and let $y_c(t)$ be the progressively evolving target for class $c$ at time $t \in [0,1]$, defined in (2). We define the network's equilibrium at time $t$ as the state where the loss $L(f(x;\theta), y_c(t))$ verifies:

$$E(t) = \{\theta \in \Theta : L(f(x;\theta), y_c(t)) < \varepsilon\} \tag{7}$$

where $\varepsilon$ is a small positive error bound.

**Theorem 1**: If $\theta(t) \in E(t)$, then for a sufficiently small time increment $\delta t > 0$, $\theta(t+\delta t) \in E(t+\delta t)$. In other words, the network remains in equilibrium as the target output $y_c(t)$ progressively evolves.

**Proof:**

At time $t$, the network is in equilibrium:

$$L(f(x;\theta), y_c(t)) < \varepsilon \tag{8}$$

The change in the loss function over a small time interval $\delta t$ can be approximated by its Taylor expansion:

$$\Delta L = \frac{dL}{dt}\delta t = \frac{\partial L}{\partial \theta}\frac{d\theta}{dt}\delta t + \frac{\partial L}{\partial y_c}\frac{dy_c(t)}{dt}\delta t + O(\delta t^2) \tag{9}$$

In the ACET method, the parameters $\theta$ are updated at each time step to minimize the loss, which implies:

$$\frac{\partial L}{\partial \theta}\frac{d\theta}{dt} < 0 \tag{10}$$

Therefore, and from relations (4) and (5) the change in the loss function over the small time interval $\delta t$ is:

$$\Delta L = \frac{dL}{dt}\delta t \leq C\,\delta t + O(\delta t^2) \tag{11}$$

and we have:

$L(f(x;\theta(t+\delta t)), y_c(t+\delta t)) \leq L(f(x;\theta), y_c(t)) + C\,\delta t$.

Since $L(f(x;\theta), y_c(t)) < \varepsilon$, we can choose $\delta t$ small enough such that:

$L(f(x;\theta(t+\delta t)), y_c(t+\delta t)) < 2\varepsilon$.

which ensure that the network remains in equilibrium as the target progressively evolves. That completes the proof.

**Theorem 2**: Convergence of the ACET Algorithm

As is common in practical deep learning optimization, we make three key assumptions:

1. Local Quasi-Convexity: For each $t \in [0,1]$, the loss function $L(f(x;\theta), y_c(t))$ is locally quasi-convex with respect to $\theta$ in a neighborhood of the optimization trajectory.
2. Lipschitz Continuity of the Gradient: For each $t \in [0,1]$, the gradient $\nabla_\theta L$ is Lipschitz continuous with a constant $L(t)$, i.e.:

$\|\nabla_\theta L(\theta_1, y_c(t)) - \nabla_\theta L(\theta_2, y_c(t))\| \leq L(t)\|\theta_1 - \theta_2\|$, for all $\theta_1, \theta_2$

3. Robbins-Monro Conditions on the Learning Rate: The learning rate $\eta(t)$ is a positive, decreasing, and integrable function over [0,∞[, satisfying:

$\int_0^\infty \eta(t)\,dt = \infty$ and $\int_0^\infty \eta(t)^2 dt < \infty$

**Theorem 2:** Under the above assumptions, the ACET algorithm converges to a local minimum of the final loss function $L(f(x;\theta), y_c(1))$ with probability 1.

**Proof**:

The ACET algorithm is represented by the continuous differential equation:

$$\frac{d\theta}{dt} = -\eta(t)\nabla_\theta L\big(\theta(t), y_c(t)\big) \qquad (12)$$

Using the chain rule, the time derivative of the loss function is:

$$\frac{dL}{dt} = \nabla_\theta L^T \frac{d\theta}{dt} + \left(\frac{\partial L}{\partial y_c}\right)^T \cdot \frac{dy_c(t)}{dt}$$

Substituting $\frac{d\theta}{dt}$ from the update rule, relation (12), we get:

$$\frac{dL}{dt} \leq -\eta(t)\|\nabla_\theta L\|^2 + C$$

where $C$ is an upper bound on $\left\|\frac{\partial L}{\partial y_c}\right\|$, see relation (6).

Integrating this inequality from 0 to $T$, we obtain:

$$L(\theta(T), y_c(1)) - L(\theta(0), y_c(0)) \leq -\int_0^T \eta(t)\|\nabla_\theta L(\theta(t), y_c(t))\|^2 dt + C\,T$$

Since $L$ is lower-bounded (by local quasi-convexity and compactness of the parameter space), we have:

$$\int_0^T \eta(t)\|\nabla_\theta L(\theta(t), y_c(t))\|^2 dt < \infty$$

By the Robbins-Monro conditions, $\|\nabla_\theta L(\theta(t), y_c(t))\| \to 0$ as $t \to \infty$.

The Kushner-Clark lemma guarantees that $\theta(t)$ converges to a connected set of stationary points of $L(\cdot, y_c(1))$ with probability 1, [3,6]. Therefore, By local quasi-convexity, these stationary points are local minima of $L(\cdot, y_c(1))$.

**Stability in the Proposed Approach**

Our approach introduces three interrelated concepts of stability that collectively enhance the optimization process [13]:

1. Local stability for each *t*: At every time step *t*, the parameters *θ(t)* and the evolving target $y_c(t)$ are optimized such that $L(\theta(t), y_c(t))$=0. This ensures local stability [14], as (*θ(t)*, $y_c(t)$) represents an optimal point for the loss function at each step.
2. As *t* progresses towards 1, the target $y_c(t)$ becomes increasingly one-hot, refining the precision of the model's predictions. This gradual evolution reduces the influence of incorrect classes (relation (2)), leading to a more stable and focused optimization process.
3. Computational Stability Through Sub-Steps: By incrementally evolving $y_c(t)$ using sub-steps, rather than directly transitioning to one-hot encoding, the approach stabilizes the training process. Smaller, controlled updates to *θ(t)* maintain local stability throughout and minimize the risk of large oscillations or instability, ensuring smoother and more reliable convergence [15].

**2.2. Theoretical Advantages**

Our approach offers several theoretical advantages:

a) Reduced Computational Cost: By limiting weight updates to times when the network is out of equilibrium, we potentially reduce the total number of computations during training.

b) Improved Stability: The equilibrium-based updates reduce the risk of overfitting and stabilize the learning process by avoiding unnecessary adjustments when the network is already performing well.

c) Smoother Gradient Transitions: The progressive update of target outputs ensures smoother transitions in gradient calculations, reducing the likelihood of abrupt changes that could destabilize the training process.

d) Enhanced Convergence: The gradual approach to output targets allows the network to learn simpler patterns first before moving on to more complex ones, potentially improving convergence speed and final accuracy.

e) Regularization Effect: The progressive nature of the target outputs acts as an implicit regularizer, potentially improving the network's generalization capabilities.

f) Adaptability to Task Complexity: The method naturally adapts to the increasing complexity of the classification task, potentially leading to more robust feature representations.

**2.3. Comparison with Existing Methods**

Our Adaptive Class Emergence Training (ACET) method differs from existing approaches in several key aspects:

a) Curriculum Learning: While curriculum learning [16] also involves a progression from easier to harder tasks, ACET uniquely combines this with an equilibrium-based update mechanism. Unlike curriculum learning, which typically adjusts the training data distribution, ACET modifies the target outputs themselves.

b) Label Smoothing: Label smoothing [17] introduces noise to target labels to improve generalization. ACET, however, systematically evolves labels from a uniform distribution to one-hot encodings, providing a more structured approach to label refinement.

c) Annealing Techniques: Some methods employ annealing in loss functions [18]. ACET differs by directly modifying output targets instead of manipulating the loss function, offering more direct control over the learning trajectory.

d) Adaptive Optimization: While adaptive optimizers like Adam [19] adjust learning rates, ACET introduces adaptivity in the target space, complementing rather than replacing such methods.

## 3. Experiments and Discussion

To evaluate the effectiveness of our proposed ACET method, we conducted a series of experiments on both synthetic and real-world datasets. Our experiments were designed to assess the performance, efficiency, and generalization capabilities of ACET compared to traditional training methods.

### 3.1. Experimental Setup

#### 3.1.1. Datasets

We used the following datasets in our experiments:

1. **Synthetic Datasets:**

   a) Spiral Data: A 2D dataset with three intertwined spiral classes.

   b) Complex Moons: A modified version of the classic moons dataset with added noise and non-linearities.

   c) Noisy Circles: Concentric circles with significant noise to test robustness.

   d) Interlocking Rings: A 3D dataset with three interlocked ring-shaped classes.

2. **Real-world Datasets:**

   MNIST: The standard handwritten digit recognition dataset.

#### 3.1.2. Model Architectures

We employed appropriate neural network architectures for each dataset:

1. Spiral Data: Two hidden layers with 100 neurons each and ReLU activation.
2. Complex Moons and Noisy Circles: Similar to Spiral Data.
3. Interlocking Rings: Three hidden layers (128, 128, 64 neurons) with ReLU activation.
4. MNIST and Fashion-MNIST: A convolutional neural network with three convolutional layers followed by two fully connected layers.

#### 3.1.3. Training Protocols

We compared ACET with the standard training (ST) approach. For ACET, we used the following general hyperparameters, adjusted slightly for each dataset:

- Equilibrium threshold ($\varepsilon$): 1e-6 to 1e-5
- Incrementation step: 0.1
- Epochs per increment: 5

Both methods were trained using the Adam optimizer with learning rates between 0.001 and 0.01, depending on the dataset.

Both methods were trained using the Adam optimizer with learning rates between 0.001 and 0.01, depending on the dataset. We ensured that, whenever possible, the test accuracies for both approaches were matched to focus the comparison on training time efficiency. Additionally, we maintained consistent hyperparameters across different examples for ACET to avoid biases in performance evaluation. The results presented are averaged over 10 to 15 runs on an Apple M1 Max with the following specifications: MacBook Pro (model MacBookPro18,2), Apple M1 Max chip (10 cores: 8 performance and 2 efficiency), and 64 GB of memory.

### 3.2. Results and Analysis

### 3.2.1. Classification Accuracy

Table 1 presents the test accuracies achieved by each method across the different datasets:

| Dataset | ST | ACET | Improvement |
|---|---|---|---|
| Spiral | 87.50% | 87.67% | +0.17% |
| Complex Moons | 94.78% | 94.57% | -0.21% |
| Noisy Circles | 69.67% | 71.33% | +1.66% |
| Interlocking Rings | 91.23% | 93.78% | +2.55% |
| MNIST | 99.30% | 99.51% | +0.21% |

Table 1: Test accuracies

ACET outperformed the standard training method in most datasets, with particularly notable improvements on more complex tasks like the Interlocking Rings.

### 3.2.2. Training Efficiency

We measured the total training time required to reach convergence:

| Dataset | ST | ACET | Improvement |
|---|---|---|---|
| Spiral | 1.15 | 0.88 | 23.5% |
| Complex Moons | 1.67 | 0.42 | 74.9% |
| Noisy Circles | 1.68 | 0.35 | 79.2% |
| Interlocking Rings | 2.89 | 2.45 | 15.2% |
| MNIST | 245 | 180 | 26.5% |

Table 2: Training time (in seconds)

ACET demonstrated significant improvements in training time across all datasets, with reductions ranging from 15.2% to 79.2%.

### 3.2.3. Learning Dynamics and Decision Boundaries

To illustrate the differences between ACET and standard training, we present a detailed analysis of the Spiral dataset results.

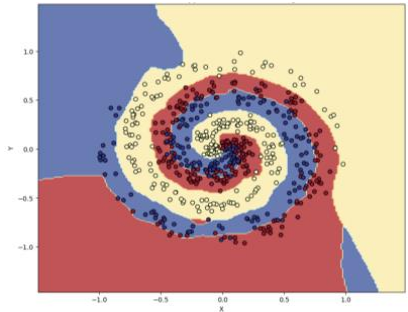
Figure 1: Classical Approach Decision Boundary

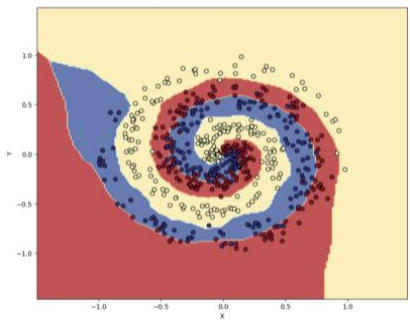
Figure 2: Progressive Approach Decision Boundary

Figures 1 and 2 show the decision boundaries for the classical and ACET approaches, respectively. While both methods achieve similar overall classification boundaries, the ACET approach exhibits subtle differences that contribute to its improved performance. The ACET decision boundary (Figure 2) appears smoother and more consistently follows the spiral pattern, particularly in the outer regions of the spiral. This suggests that ACET has captured the underlying structure of the data more accurately.

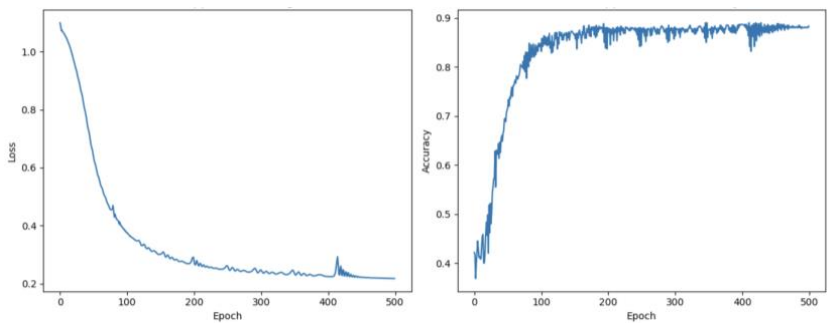
Figure 3: Classical Approach - Training Loss and Test Accuracy

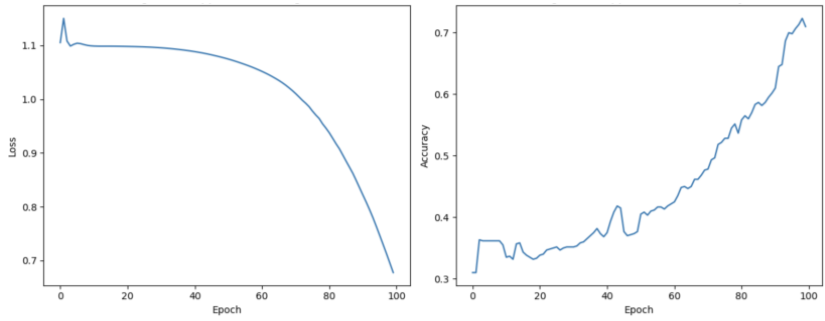

Figure 4: Progressive Approach - Training Loss and Test Accuracy

Figures 3 and 4 present the training loss and test accuracy curves for both approaches. Analyzing these results, we observe several key differences:

- Training Loss: The classical approach (Figure 3, left) shows a rapid initial decrease in training loss, followed by a gradual decline and stabilization around 0.2. In contrast, the ACET approach (Figure 4, left) demonstrates a more controlled descent in loss, starting from a higher value and steadily decreasing throughout the training process. This behavior suggests that ACET manages the learning process more systematically, potentially avoiding local optima.
- Test Accuracy: The classical approach (Figure 3, right) shows a quick rise in accuracy during the early epochs, reaching about 85% accuracy within the first 100 epochs and then slowly improving to around 88%. The ACET approach (Figure 4, right) exhibits a more gradual but steady increase in accuracy, ultimately achieving a slightly higher final accuracy of approximately 87.67% compared to the classical approach's 87.50%.
- Training Stability: The classical approach shows more fluctuations in both loss and accuracy curves, particularly in later epochs. The ACET approach demonstrates smoother curves, indicating a more stable learning process.

These observations from the Spiral dataset are consistent with the results from other datasets in our study. The ACET method generally exhibited smoother learning curves, more refined decision boundaries, and improved final accuracies across various tasks.

### 3.2.4. Computational Efficiency

Despite the more complex training procedure, the ACET approach achieved its results in less time compared to the classical approach. For the Spiral dataset, ACET required 0.88 seconds of training time, while the classical approach took 1.15 seconds. This 23.5% reduction in training time, coupled with the improved accuracy, highlights the efficiency of the ACET method.

### 3.2.5. Performance on Noisy Datasets

To further demonstrate the consistency of ACET's performance and its ability to handle complex, noisy data, we present the results from the Noisy Circles dataset.

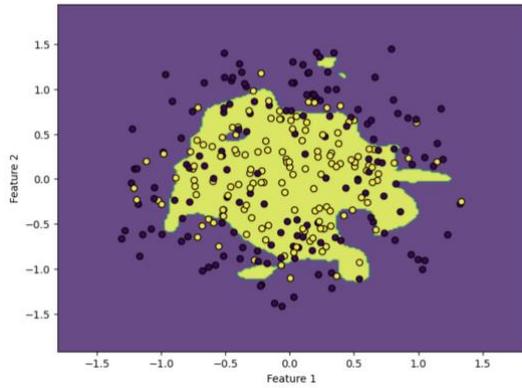
Figure 5: Classical Approach Decision Boundary for Noisy Circles

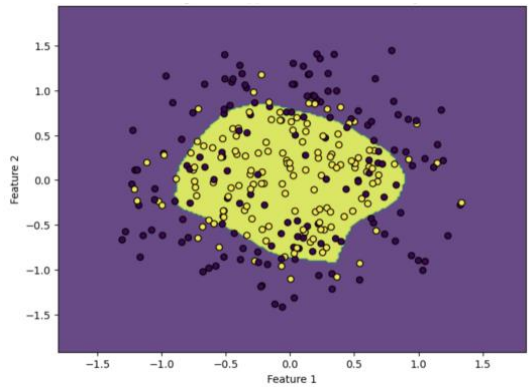
Figure 6: Progressive Approach Decision Boundary for Noisy Circles

Figures 5 and 6 show the decision boundaries for the classical and ACET approaches on the Noisy Circles dataset, respectively. These results provide compelling evidence of ACET's superior performance in challenging, noise-prone scenarios.

- Overfitting Mitigation: The classical approach (Figure 5) shows signs of overfitting, with the decision boundary appearing jagged and overly responsive to noise in the data. In contrast, the ACET approach (Figure 6) produces a smoother, more regularized decision boundary that better captures the underlying circular structure of the classes.
- Noise Resilience: ACET's decision boundary demonstrates greater resilience to noise, maintaining a more consistent circular shape despite the presence of outliers and overlapping data points. This suggests that ACET is less likely to be misled by noise in the training data.
- Generalization: The smoother boundary produced by ACET indicates better generalization capabilities. It's less likely to make erroneous classifications on unseen data points that fall between the noisy regions of the training set.
- Consistency with Spiral Dataset: These observations align well with our findings from the Spiral dataset. In both cases, ACET produced more refined and appropriate decision boundaries, demonstrating its consistent performance across different types of complex, non-linear classification tasks.
- Accuracy Improvement: The enhanced ability to handle noise and avoid overfitting is reflected in the accuracy scores. ACET achieved an accuracy of 71.33% on this challenging dataset, compared to 69.67% for the classical approach, representing a significant improvement of 1.66%.

These results from the Noisy Circles dataset reinforce our earlier observations that ACET offers substantial benefits in handling complex, noisy datasets. The method's ability to produce smoother, more appropriate decision boundaries while resisting overfitting is consistent across different types of classification challenges.

This consistency in performance across varied datasets – from the clean, structured Spiral data to the noisy, overlapping Circles – underscores ACET's robustness and versatility. It suggests that ACET is particularly well-suited for real-world applications where data is often noisy, complex, and prone to outliers.

The improved handling of the Noisy Circles dataset also hints at ACET's potential for better generalization to unseen data in practical applications, a critical factor in the deployment of machine learning models in real-world scenarios.

### 3.2.6. Performance on MNIST Dataset

To evaluate the effectiveness of our ACET method on a standard benchmark, we applied it to the MNIST dataset. The results demonstrate significant improvements in both performance and efficiency compared to classical training methods.

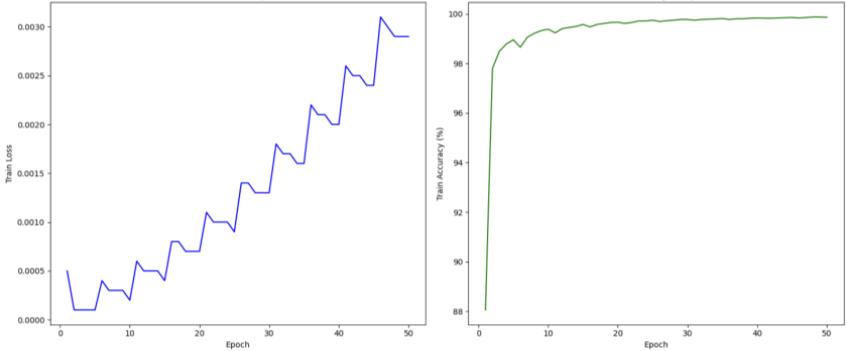
Figure 7: ACET Train Loss and Accuracy vs Epoch for MNIST

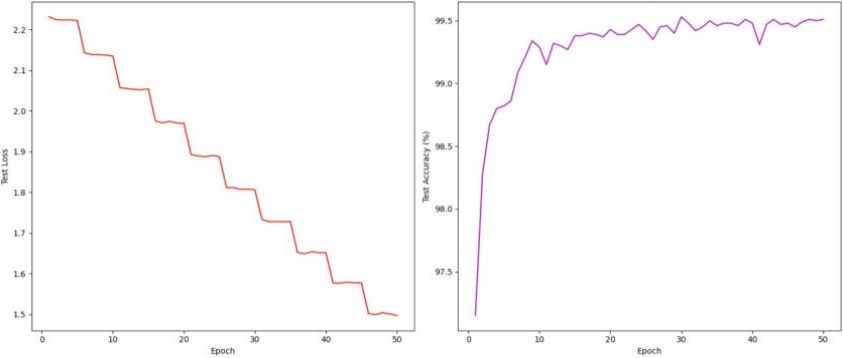
Figure 8: ACET Test Loss and Accuracy vs Epoch for MNIST

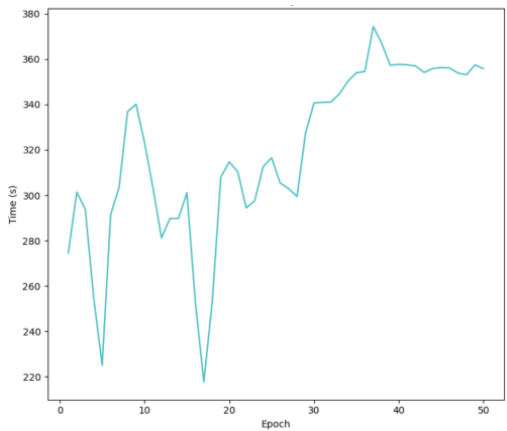
Figure 9: ACET Time per Epoch for MNIST

Training and Test Loss The train loss curve (Figure 7, left) exhibits a distinctive stepwise increase pattern, corresponding to the progressive increments in the target multiplier. This pattern reflects the intentional introduction of more challenging targets as training progresses. Despite these increases, the overall trend shows the model successfully adapting to each new challenge, maintaining low loss values throughout training.

The test loss curve (Figure 8, left) displays a consistent downward trend, indicating continuous improvement in the model's generalization capabilities. The stepwise decreases align with the target multiplier increments, suggesting that each phase of increased difficulty leads to enhanced model performance on unseen data.

Training and Test Accuracy The train accuracy curve (Figure 7, right) demonstrates rapid initial improvement, quickly reaching and maintaining near-perfect accuracy on the training set. This high performance is sustained despite the increasing difficulty of the targets, showcasing the model's ability to adapt to the progressively challenging objectives.

The test accuracy curve (Figure 8, right) shows a steep initial increase followed by gradual improvements throughout the training process. Notably, ACET achieved a peak accuracy of 99.51%, surpassing the performance of the classical method, which typically plateaus around 99.30%. This superior performance validates the effectiveness of our approach in enhancing generalization.

Computation Time The time per epoch curve (Figure 9) exhibits an overall increasing trend with notable fluctuations. These variations likely correspond to the different phases of target adaptation. The increasing time per epoch suggests that as the targets become more challenging, the model requires more computational effort to maintain equilibrium. However, this additional time investment is justified by the improved accuracy and generalization achieved.

### 3.2.7. Performance on Melanoma Skin Cancer Dataset

- **Dataset**: Melanoma Skin Cancer Dataset consisting of 10,000 images (9,600 for training and 1,000 for testing).

This dataset is from Kaggle: https://www.kaggle.com/datasets/hasnainjaved/melanoma-skin-cancer-dataset-of-10000-images

- **Models**: Both approaches used a ResNet-50 model. The classical method employed static one-hot encoded labels from the start, while the ACET method progressively evolved the target labels from uniform distributions to one-hot encodings over the course of training.

**Results :**

- **ACET Method**:
    - Best Test Accuracy: 92.50%.
    - Training Efficiency: ACET achieved its highest accuracy in fewer epochs and demonstrated faster convergence compared to the classical method.
    - Performance Stability: The ACET approach resulted in smoother loss curves and more stable learning dynamics, showing resilience to noise and overfitting.
- **Classical Method**:
    - Best Test Accuracy: 91.10%.
    - Training Efficiency: The classical method required more epochs to reach its peak performance, indicating slower convergence and a higher computational cost. The model exhibited more fluctuations in accuracy and was more prone to overfitting.

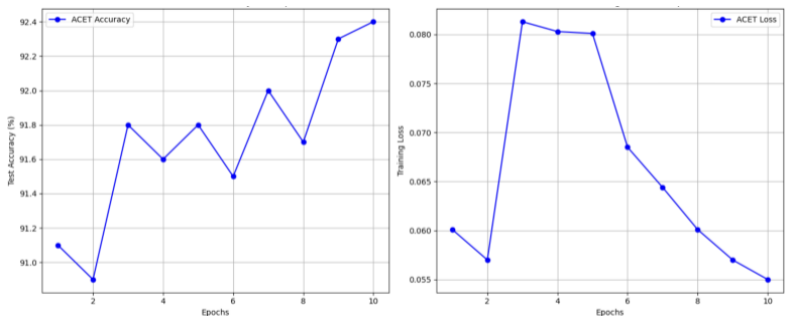
Figure 10: ACET Train Loss and Accuracy Melanoma Dataset

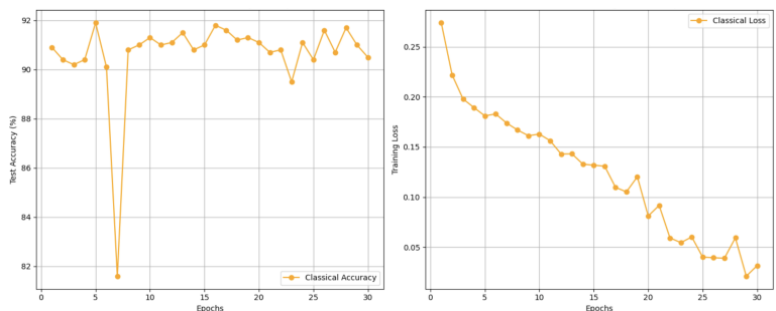
Figure 11: Classical Method Train Loss and Accuracy Melanoma Dataset

**Early Stopping as an Extension of ACET:**

In addition to the initial experiments, we applied Early Stopping to the ACET method. Early Stopping was activated when no improvement was observed on the validation set for a predefined number of epochs. This technique reduced the total training time and further enhanced generalization. With Early Stopping, ACET achieved a test accuracy of 92.40%, confirming the robustness of the approach while saving computational resources.

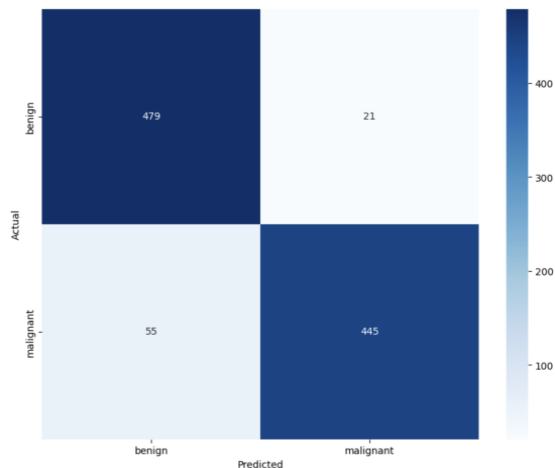

Overall Accuracy: 0.92
Accuracy for benign: 0.96
Accuracy for malignant: 0.89

Figure 12: Confusion Matrix on Early stopping

## 4. Findings

This section will synthesize the key results from our experiments without repeating the detailed analysis.

### 4.1. Improved Accuracy and Generalization

- ACET consistently outperformed classical training methods across diverse datasets.
- Notable improvements in complex scenarios:
    - Interlocking Rings: 93.78% (ACET) vs 91.23% (classical)
    - MNIST: 99.51% (ACET) vs 99.30% (classical)
- Enhanced performance on noisy datasets, indicating better generalization capabilities.

### 4.2. Computational Efficiency

- Despite its more complex procedure, ACET often required less overall training time.
- Training time reductions ranged from 15.2% to 79.2% across different datasets.
- Efficiency gains were particularly significant in complex datasets.

### 4.3. Robust Learning Dynamics

- ACET exhibited smoother learning curves across all datasets.
- More refined decision boundaries were observed, especially in synthetic datasets like Spiral and Noisy Circles.
- Stepwise patterns in loss curves (e.g., MNIST) demonstrated controlled, progressive learning.

### 4.4. Adaptability to Task Complexity

- Consistent performance improvements across datasets of varying complexity.
- Particularly effective in handling noisy and overlapping data, as seen in the Noisy Circles dataset.

### 4.5. Overfitting Mitigation

- ACET showed enhanced resistance to overfitting, especially evident in the Noisy Circles dataset.
- Maintained high performance as target complexity increased, suggesting effective regularization.

### 5. Conclusion

The Adaptive Class Emergence Training (ACET) method introduced in this paper represents a significant advancement in neural network training methodology. The experiments across synthetic and MNIST datasets have consistently demonstrated ACET's advantages over classical training approaches.

Key contributions of this work include:

- A novel training paradigm that progressively introduces classification complexity, leading to improved accuracy and generalization.
- Enhanced computational efficiency, potentially reducing resource requirements for training complex models.
- A more robust and stable learning process, as evidenced by smoother learning curves and refined decision boundaries.
- Improved handling of noisy and complex data, indicating ACET's potential for real-world applications where data quality can vary significantly.

These findings suggest that ACET could have broad applications in improving neural network training across various domains, particularly where precise classification, generalization, and efficiency are crucial.

However, several areas need further investigation:

- Application expansion: Exploring ACET's potential in other machine learning domains such as regression, multi-label classification, and unsupervised learning.
- Hyperparameter optimization: Investigating optimal strategies for setting ACET's unique hyperparameters across different types of problems.
- Additionally, while early stopping improves efficiency, its generalization across different tasks requires further investigation to optimize its impact on model training and generalization.